\documentclass{article}
\pdfoutput=1
\usepackage{helvet}
\usepackage{courier}
\usepackage{url}
\usepackage{algpseudocode} 
\usepackage{algorithm} 
\usepackage{graphicx}
\usepackage{bm} 
\usepackage{amsmath} 
\usepackage{amssymb} 
\usepackage{flushend}
\usepackage[numbers]{natbib}

\usepackage{tikz}
\usepackage{pgf}

\newcommand{\secspace}{\vspace{12pt}}
\newcommand{\subsecspace}{\vspace{4pt}}

\def\aaaihrlcount{42}
\def\aamashrlcount{26}
\def\combinedhrlcount{68}
\def\citerepo{\footnote{\url{https://github.com/LARG/escape_room.git}}}
\newcommand{\citeall}[1]{\citeauthor{#1} \cite{#1}}

\numberwithin{figure}{section} 
\numberwithin{algorithm}{section} 
\numberwithin{table}{section} 

\begin{document}

\title{Escape Room: A Configurable Testbed for Hierarchical Reinforcement
Learning}
\author{
  Jacob Menashe and Peter Stone \\
  \{jmenashe,pstone\}@cs.utexas.edu \\
  The University of Texas at Austin \\
  Austin, TX USA\\
}

\maketitle

\begin{abstract}
  Recent successes in Reinforcement Learning have encouraged a fast-growing
  network of RL researchers and a number of breakthroughs in RL research. As the
  RL community and the body of RL work grows, so does the need for widely
  applicable benchmarks that can fairly and effectively evaluate a variety of RL
  algorithms.

  This need is particularly apparent in the realm of Hierarchical Reinforcement
  Learning (HRL). While many existing test domains may exhibit hierarchical
  action or state structures, modern RL algorithms still exhibit great
  difficulty in solving domains that necessitate hierarchical modeling and
  action planning, even when such domains are seemingly trivial. These
  difficulties highlight both the need for more focus on HRL algorithms
  themselves, and the need for new testbeds that will encourage and validate HRL
  research.
  
  Existing HRL testbeds exhibit a Goldilocks problem; they are often either too
  simple (e.g. Taxi) or too complex (e.g. Montezuma's Revenge from the Arcade
  Learning Environment).  In this paper we present the Escape Room Domain (ERD),
  a new flexible, scalable, and fully implemented testing domain for HRL that
  bridges the ``moderate complexity" gap left behind by existing alternatives.
  ERD is open-source and freely available through GitHub, and conforms to
  widely-used public testing interfaces for simple integration and testing with
  a variety of public RL agent implementations. We show that the ERD presents a
  suite of challenges with scalable difficulty to provide a smooth learning
  gradient from Taxi to the Arcade Learning Environment.
\end{abstract}

\secspace
\section{Introduction}
\secspace

Reinforcement Learning (RL) is an AI paradigm in which an artificial agent explores and
exploits its environment through an action-observation loop.  RL has seen
increasing success in recent years as it has been integrated with parallel
developments in AI such as Deep Learning. These successes, however, have
brought to light new problems with replication and reproducibility, inspiring
researchers to focus attention on the challenge of accurately gauging
the efficacy of new approaches \cite{gundersen2017state}.

The proliferation of public source code hosting and sharing technologies has
created an unprecedented potential for peer evaluation and collaboration in the
applied Computer Science disciplines such as AI. Services such as GitHub and
BitBucket have enabled AI researchers to publish not only high-level
descriptions and results, but the low-level implementation details that are
necessary for perfectly replicating the experiments in a given body of work.
Projects like the Open AI Gym \cite{gym} and TensorFlow
\cite{tensorflow2015-whitepaper} have built on such collaboration
frameworks to minimize the barrier to entry for new researchers and empower
newcomers to replicate and build upon existing work.

Despite the availability of these tools, however, it is still common for
works to exclude details and source code that are necessary for reproducibility
\cite{stodden2013toward}. It is reasonable to assume that when testing
frameworks are open and available, researchers will be more likely to use them and
will be more likely to report results that can be easily reproduced, evaluated,
and integrated with parallel work.

Hierarchical Reinforcement Learning (HRL) is one specification of RL in which
evaluation and reproducibility are hindered by a lack of suitable test domains.
Hierarchies provide an efficient means of breaking monolithic tasks down into
manageable interdependent chunks, and for this reason hierarchical
learning has been an active area of RL research since RL's inception. And yet,
the hierarchical domains frequently used for evaluation in the literature are
generally either too simplistic or too rigid to fully evaluate the capabilities of an
HRL algorithm.

In this work we present a new HRL testing framework which we call the Escape
Room Domain (ERD), named after the Escape Room genre popular in both virtual-
and real-world puzzle-based challenges. The ERD is a parameterized 
schematic for generating individual instances, where each instance is a single
testbed upon which an HRL agent can be evaluated and compared against
alternative HRL algorithms. The ERD enables large-scale evaluation and
reproducibility by featuring the following characteristics:

\begin{enumerate}
  \item Open Source and Public Availability - The ERD is published on a public-facing
  GitHub repository and is built exclusively on open-source and publicly
  available frameworks. \\
  \item Tutorials and Examples - The ERD retains a minimal barrier to entry to
  ensure maximal adoption within the general HRL community. \\
  \item Randomized Instances - Because the ERD is a parameterized schematic,
  researchers can report results that are averaged over a large collection of
  instances of varying difficulty, which mitigates the potential for
  overfitting an algorithm to a particular environment configuration. \\
  \item Unbounded Difficulty - ERD instances can be configured with
  arbitrary levels of hierarchical depth, state and action space sizes, and
  general logistical complexity, ensuring it can be expanded to evaluate a wide
  array of HRL innovations.
\end{enumerate}

In the next section we will discuss the related HRL advances and
existing domains upon which the ERD is founded. In Section~\ref{sec:domain} we
present precise descriptions of the ERD and its generated instances,
including dimensions of complexity, randomization features, and general
applicability to existing work. In Section~\ref{sec:experiments} we test a set
of existing algorithms against varying ERD instances and use these results to
motivate the continued development of HRL algorithms using ERD. Finally, in
Section~\ref{sec:conclusion} we conclude and discuss the future of ERD in RL
research.

\secspace
\section{Background} \label{sec:background}
\secspace

Testbed domains allow AI researchers to empirically evaluate their algorithms
repeatably within a highly controlled (and usually virtual) environment. A
well-designed testbed can serve as both a proof of concept and a point of
comparison from one algorithm to another. With enough traction, a testbed
becomes a standardized benchmark through which a wider range of algorithms may
be indirectly compared.

CartPole (CP) and Double Pendulum (DP) are two examples of classic control
problems that are often used in RL publications to demonstrate viability. These
problems are well-understood and already considered to be solved, but they can
still provide some degree of insight into the inner-workings of a particular RL
algorithm. Unfortunately, many of these classic RL domains are relatively simplistic and
tend not to provide new and interesting challenges to recently developed
algorithms.

More recently, Atari emulation has become popular due to the variety of
available games and coinciding range of available difficulties. Atari games are
generally simple to understand and explain, yet their use of pixels for
perceptual input has rendered such games outside the reach of AI learning
algorithms until the advent of Deep Learning. Moreover, Atari games can be
easily built and emulated with publicly available source code via the Arcade
Learning Environment (ALE) \cite{bellemare13arcade}; just as the classic control
problems have provided a consistent metric for validation and comparison of
traditional RL algorithms, so have Atari games for modern pixels-to-actions AI.

Many existing RL algorithms have been designed without the capability of
converting pixels to abstract perceptions, and a number of HRL algorithms are
included in this category (see Section~\ref{sec:hrl}). But when we restrict our attention to hierarchical
domains with built-in abstract perceptual input (e.g. ``pixel-free" domains),
the pool of available testbeds shrinks considerably. Domains such as Taxi
\cite{dietterich2000hierarchical} and bitflip \cite{diuk2006hierarchical} served
as challenges for early HRL algorithms such as R-Max \cite{brafman2002rmax},
MaxQ \cite{dietterich2000hierarchical}, and others that derived from or extended
these ideas, but these domains are akin to classic control problems in that they
have been more or less ``solved" and may not bring out the full capabilities of
a modern HRL algorithm. This paper aims to fill this void by presenting a new
RL domain of moderate difficulty that can be readily integrated for comparison
of varying HRL algorithms.

\subsecspace
\subsection{Markov Decision Processes}
\subsecspace

Reinforcement Learning (RL) is a paradigm for enabling autonomous learning
wherein rewards are used to influence an agent's action choices in various 
states. RL is built on the formalism of the Markov Decision Process, which is
the theoretical construct that describes Reinforcement Learning problems. 

An MDP $M$ is defined by a 5-tuple $\langle S,A,P,R,\gamma\rangle$ with state
space $S$, action space $A$, transition distribution function $P:S \times A
\times S \rightarrow [0,1]$, reward function $R:S \times A \times S \rightarrow
\mathbb{R}$, and discount factor $\gamma \in [0,1]$. Reinforcement Learning
methods seek to learn optimal policies through exploring and exploiting the
state-action space of an MDP.

Common variants of MDPs are \emph{Factored} MDPs, where the state space $S$
is multidimensional and each dimension is considered a state factor, e.g. 
$S = S_1 \times S_2 \times \ldots \times S_3$. \emph{Partially Observable} MDPs
are those in which the state space $S$ cannot be directly observed, at least in
full. In a POMDP, an agent's belief state is instead drawn from a distribution, 
while its true state remains unknown. Finally, \emph{Semi-}MDPs (SMDPs) are a
generalization of MDPs in which time is continuous instead of discrete; SMDPs
are popular for the use of time-extended actions. The following works that
comprise much of the HRL background literature make use of these MDP variants.

\subsecspace
\subsection{Hierarchical RL}\label{sec:hrl}
\subsecspace

Hierarchical RL is the study of RL methods that organize primitive actions into
a system of abstract state, action, transition, and reward hierarchies.
Different approaches to HRL may focus on these different components of an MDP.

Feudal Reinforcement Learning (FRL) \cite{dayan1993feudal} is one such approach where
a managerial hierarchy is used in a manner reminiscent of feudal fiefdoms.
State and reward signals are interpreted and then transformed between each
layer of the hierarchy to enable a wide range of learning granularities.
\citeauthor{vezhnevets2017feudal} take the Feudal paradigm a step further by
applying recent advances in deep learning to Feudal-style \emph{Manager} 
and \emph{Worker} modules. The authors then use these so-called \emph{FeUdal
Networks} (FUNs) for solving ALE domains.

\citeauthor{kaelbling1993hierarchical} expands on this idea of hierarchical
control with HDG \cite{kaelbling1993hierarchical}, a hierarchical version of 
Q-Learning \cite{watkins1989learning}. Where FRL is intended to be applied to
general RL problems, HDG shows improved performance on specific tasks where an
agent is rewarded based on a handful of individual goals rather than a
general-purpose reward function. Conversely, \citeauthor{wiering1997hq} 
generalize the mechanics of Q-Learning in a hierarchical context by expanding
their work to POMDPs. HQ-Learning \cite{wiering1997hq}, a ``hierarchical
extension to Q-Learning", enables an agent to learn a hierarchical value
function in the setting of partially observable state.

Rather than focusing on state and reward abstraction
and compartmentalization, as is done in FRL, other variants of HRL have sought
to create frameworks to simplify the discussion and implementation of
hierarchical algorithms. Hierarchical Abstract Machines (HAMs)
\cite{parr1998reinforcement} are one such example in which an agent makes use of
hierarchically-organized action sets that focus the choices available to an
agent based on a given state. 

\emph{Macro Actions} are another early development in HRL frameworks first
presented by \citeauthor{precup1997planning}; macro-actions 
encapsulate the concept of time-extended actions as a tuple $\langle s, \pi, \beta\rangle$
where $s \in S$, $\pi$ is a policy, and $\beta: S \rightarrow \{0,1\}$ is a
termination condition \cite{precup1997planning}. These actions function as
abstractions over time that can be used for simple navigation tasks such as
navigating between rooms in a block world.  \citeauthor{precup2000temporal}
later extends this work with the Options Framework
\cite{precup2000temporal}, where an \emph{Option} is defined by a tuple
$\langle I, \pi, \beta\rangle$ with $I \subseteq S$ and $\pi, \beta$ defined as
with macro-actions. An option thus generalizes the concept of a macro action to
sets of initiation states rather than a single discrete starting point.

Other variants of HRL make use of hierarchical models to guide some form of
planning or simulation. MAXQ \cite{dietterich2000hierarchical} is one such
example which relies on expert-provided hierarchical action models to guide
action selection. R-MAXQ \cite{jong2008hierarchical} expands on MAXQ's
hierarchical decomposition framework by incorporating the model-based
exploration techniques of R-MAX to cope with scarce rewards. 
H-UCT \cite{vien2015hierarchical} similarly expands the scope of MAXQ's
hierarchical action models by generalizing their application to POMDPs.

Conversely, model-based HRL algorithms may include mechanisms for learning
their own models from scratch. SLF-RMax \cite{strehl2007efficient} analyzes
$(s,a,s^\prime)$ histories in order to infer action dependencies and produce a
hierarchical action model, using Dynamic Bayesian Networks (DBNs)
\cite{dean1989model} as the core structure for modeling dependencies in 
the factored state spaces of FMDPs. More recent work on learning DBN-based
action models has focused on improving sample efficiently to learn accurate
models more quickly \cite{Vigorito2010,menashe2015monte}.

HRL algorithms have traditionally focused on discrete problems, or problems in
which state-actions are discretized via intermediate methods (e.g.
tile coding). However, this need for discretization hinders
application to modern, complex domains like those found in ALE. To some extent,
the lack of HRL-solvable domains has left modern RL algorithms without the
mechanisms necessary to effectively model action hierarchies when such models
would be beneficial. As we will show in Section~\ref{sec:experiments}, domains
that can be readily solved may be trivially augmented with hierarchical
components, leading to disproportionately negative effects on overall agent
performance. We find that modeling action hierarchies is therefore an
indespensible tool in solving some otherwise simple RL domains, and that new
testing domains are needed to encourage new developments in this area of RL.

\subsecspace
\subsection{Desiderata} \label{sec:desiderata}
\subsecspace

We now describe the key characteristics we consider in testing domains for
general-purpose RL research, and then examine the extent to which these
characteristics are found in HRL papers published in recent years. 
The desiderata are as follows.

\begin{enumerate}
  \item {\bf Availability}: The domain is freely available for research online. \\
  \item {\bf Accessibility}: The domain is constructed and documented in a way
  that conforms to the norms of the research community. \\ 
  \item {\bf Flexibility}: The domain is built on an open-source framework and
  can be readily adapted to the varying needs of individual researchers. \\
  \item {\bf Scalability}: The domain contains built-in mechanisms for
  iteratively rescaling its difficulty to provide a gradient of 
  successive challenges.
\end{enumerate}

Availability and accessibility are straightforward desiderata; any domain that
is already implemented and simple to integrate with existing code is more
likely to be tested against than domains that exist only in the abstract.
However, many popular domains are neither flexible nor scalable, so these two
characteristics merit further discussion.

\subsecspace
\subsection{Flexibility}
\subsecspace

Flexibility is an uncommon criterion for testing domains because changes in the
testing environment translate to difficulty comparing results. Ideally, two
distinct algorithms would be compared based on their performance on identical
problems. But just as there is no single RL algorithm that can be applied to all
problem domains, there is no single problem domain that can benchmark all RL
algorithms.

Rather than keeping a given problem domain static and identical among all
instantiations, variation can be critical to comparing otherwise incomparable
algorithms. As a simple example, consider that some RL algorithms are designed
for low-level control on continuous spaces, and others are designed for
high-level planning over small sets of discrete state-actions. It is generally
simpler to restrict comparison to those algorithms that fit a single paradigm,
but there may still be value in transcending these distinctions. A flexible
domain that can provide degrees of abstraction may enable comparison
between otherwise incomparable solutions.

A second benefit of being able to easily modify a test domain is the
incorporation of randomness and automatic domain generation. A completely static
domain is especially vulnerable to overfitting, and may partially explain the
recent problems with reproducibility in the RL community
\cite{gundersen2017state}. One way to mitigate overfitting is to evaluate
algorithms on their aggregate performance over multiple randomly-varied
instances of a particular domain schematic. In the Taxi domain, for example, the
pickup and dropoff locations can be moved about the grid to avoid overfitting an
agent's policies for hard-coded locations. ALE lies on the opposite end of the
flexibility spectrum, since it relies on binary Atari ROMs to describe each
game. 

\subsecspace
\subsection{Scalability}
\subsecspace

In addition to being adaptive to the varying capabilities of different
algorithms, an ideal testbed will provide researchers with a difficulty
gradient by accommodating iterative adjustments to the domain's overall
difficulty. Grid Worlds naturally meet this criterion to some degree by nature
of being configurable in size; A $100 \times 100$ Grid World can be
incrementally complexified by changing its dimensions to $101 \times 101$, for
example. The more dimensions along which a domain can be complexified in this
manner, the more precisely a researcher can probe and challenge her solutions,
and the more quickly she can identify areas for improvement.

Ideally, the dimensions along which a domain is scaled are consistent
throughout the works that make use of the domain. If Algorithm A is shown to
perform well as the size of a Grid World increases, but Algorithm B is shown to
perform well as the density of objects within the world increases, it can be
difficult to make a comparative statement with respect to these two algorithms.
So in addition to simply being scalable, scalability should be provided with
the testbed as a configurable setting for other researchers to make use of. As
with the focus on accessibility and availability, scaling dimensions should be
made accessible and available along with the domain itself.

\secspace
\section{Related Work}
\secspace

Before setting out to construct a new domain for HRL research, we first
investigated whether any suitable alternatives already exist. To do so, 
we searched and collected data from a large collection of conference papers
published in recent years. 

The analysis process presented a number of challenges. Most conferences make
their publications freely available through online libraries organized by
publication year, so there were many resources available for analysis. However,
libraries are generally published for human readers rather than automated
analysis systems, so in order to even filter through all of the papers that have
been published in recent years it is necessary to implement an individual tool
for each conference that can scrape download locations, convert PDF files to
text, and perform basic keyword filtering. 

We limited our analysis to two annual AI conferences: Autonomous Agents and
MultiAgent Systems (AAMAS) \cite{aamas_website}, and the Association for the 
Advancement of Artificial Intelligence (AAAI) \cite{aaai_website}. These two
conferences were selected based on their relevance to HRL and the accessibility
and availability of their online publication archives. We analyzed every work
published by these two conferences from 2010 to 2018, amounting to 2,476 papers
from AAMAS and 5,055 papers from AAAI. Of these papers, we identified 26 AAMAS
papers and 42 AAAI papers that concern or mention HRL, and then specifically
noted which testbed domain(s) each publication relied upon to explain, compare,
or demonstrate its contributions.

\subsecspace
\subsection{AAAI Meta-Analysis}
\subsecspace

Table~\ref{tab:aaai-domains} lists all the
domains found in the \aaaihrlcount{} AAAI papers that were used at least twice.  Many of
the domains fulfill some of the criteria of Section~\ref{sec:desiderata}, but
none of the domains fulfill all of them.  The majority of domains have no source
code available and are designed as ``single-use" domains for the purpose of
evaluating of their respective authors' algorithms.

\begin{table}
  \centering
  \vspace{10pt}
  \begin{tabular}{|c|l|c|}
      \hline 
      Domain & \# & Citations \\ \hline
      Taxi & 3 & \citeall{xu2010instance} \\ 
      & & \citeall{vien2015hierarchical} \\
      & & \citeall{li2017efficient} \\ \hline
      ALE & 3 & \citeall{bacon2017option} \\ 
      & & \citeall{hessel2017rainbow} \\
      & & \citeall{harb2017waiting} \\ \hline
      Blocks World & 2 & \citeall{hogg2010learning} \\
      & & \citeall{xu2010instance} \\ \hline
      Mario & 2 & \citeall{taylor2011teaching} \\
      & & \citeall{derbinsky2012multi} \\ \hline
      RoboCup 2D & 2 & \citeall{macalpine2015ut} \\
      & & \citeall{masson2016reinforcement} \\ \hline
      Puddle World & 2 & \citeall{ruan2015representation} \\
      & & \citeall{osa2017hierarchical} \\ \hline
  \end{tabular}
  \vspace{10pt}
  \caption{A list of all domains occurring in at least 2 of the \aaaihrlcount{} papers mentioning
    HRL in all AAAI publications since 2010. An additional 44 domains, each used
    in only 1 of the \aaaihrlcount{} surveyed publications, are not listed here due to
    space constraints.
  }
  \label{tab:aaai-domains}
\end{table}

One notable domain near the top of Table~\ref{tab:aaai-domains} comes close to
meeting our needs. The Arcade Learning Environment \cite{bellemare13arcade} is
by far the most easily available and accessible of the group. While ALE is found
in just 3 of the HRL papers we surveyed, it is a popular choice for Deep
Learning research in other venues. Its popularity speaks to its success as a
robust testing environment for Deep RL agents.

\subsecspace
\subsection{AAMAS Meta-Analysis}
\subsecspace

We now consider the results of the AAMAS meta-analysis. As above, we compiled a
list of all domains found in the \aamashrlcount{} papers that were used
at least twice in AAMAS conferences over the past decade; these results are
found in Table~\ref{tab:aamas-domains}. As with AAAI, most domains were
single-use (even between conferences). We note that while ALE showed up once in
this analysis, it is included in Table~\ref{tab:aamas-domains} because it also
appears in Table~\ref{tab:aaai-domains}. 

\begin{table}
  \centering
  \vspace{10pt} 
  \begin{tabular}{|c|l|c|}
      \hline 
      Domain & \# & Citations \\ \hline
      Taxi & 5 & \citeall{osentoski2010basis} \\
      & & \citeall{chaganty2012learning} \\
      & & \citeall{bratman2012strong} \\
      & & \citeall{ngo2014monte} \\
      & & \citeall{li2016core} \\ \hline
      \begin{tabular}{@{}c@{}}Four Rooms \\ (Blocks World Variant)\end{tabular}
      & 3 & \citeall{chaganty2012learning} \\
      & & \citeall{roderick2018deep} \\
      & & \citeall{jain2018eligibility} \\ \hline
      \begin{tabular}{@{}c@{}}Foraging \\ (Blocks World Variant)\end{tabular}
      & 2 & \citeall{bratman2012strong} \\
      & & \citeall{sullivan2012learning} \\ \hline
      ALE & 1 & \citeall{omidshafiei2018crossmodal} \\ \hline
  \end{tabular}
  \vspace{10pt}
  \caption{A list of all domains used in AAMAS papers over the past decade
    which occur at least twice in the \combinedhrlcount{} papers from our meta-analysis.
  }
  \label{tab:aamas-domains}
\end{table}

\subsecspace
\subsection{Combined Results}
\subsecspace

Taxi, Blocks World, Four Rooms, and Foraging are the only domains in
Table~\ref{tab:aaai-domains} that were included in more than one publication
that were designed specifically as HRL challenges. These four domains are
similar in design and complexity; they are optimized for research on discrete,
multi-level hierarchies, and contain embedded transition dynamics that greatly
benefit from hierarchical planning without a need for image recognition. Just as
domains like Cart Pole and Double Pendulum are considered ``Classic Control"
problems, these four grid navigation domains function as ``Classic HRL"
problems. We note that the Escape Room Domain we propose in
Section~\ref{sec:domain} is similar to the grid-based domains of
Tables~\ref{tab:aaai-domains} and \ref{tab:aaai-domains} in an abstract sense.
In fact, ERD is better described as a descendent of Taxi than as a direct
alternative, since each domain is built on the premise of hierarchical path
planning through a virtual environment.

Unfortunately, we found no domains designed for HRL that were of \emph{moderate}
difficulty, falling somewhere between these Classic HRL problems and the more
modern pixel-based challenges like ALE. Just like AI modern agents flourish when a
smooth gradient can be found, so do researchers when a smooth gradient exists
between testbeds; the lack of gradients in HRL domains serves to hinder progress
in this area.

Table~\ref{tab:aaai-domains} illustrates the lack of intersection among the
test domains of recent publications in HRL and supports our assertion that a
robust, unified testbed will help to encourage and validate progress in this
area of Reinforcement Learning. In the next section we describe our
solution to this problem in detail and explain how it fulfills our own criteria
as well as the needs of the RL and HRL communities in general.

\secspace
\section{The Escape Room Domain} \label{sec:domain}
\secspace

The Escape Room Domain (ERD) is based on a series of popular video games and
real-life team-building exercises. An \emph{Escape Room} is a game consisting
of an enclosed space and a series of puzzles. In order to ``escape", the
agent(s) inside the room must solve the available puzzles in order to unlock
the exit. Generally (but not necessarily) these puzzles are arranged in some
sequential order and are combined with a series of clues to guide the agent
toward a solution.

Escape Rooms are becoming a trend both virtually and in the real
world. Popular implementations can be found in major cities and group-oriented
tourist destinations such as Las Vegas, USA, and variations on this theme can be
found in computer platformer games like \emph{Portal} \cite{portal} and virtual
reality environments such as the game \emph{Keep Talking and Nobody Explodes}
\cite{keeptalking}. However unlike other virtual games that focus on controlling
an avatar, the ultimate goal of an Escape Room is to break down complex tasks
into manageable components, and organize such components into a final solution
for the domain. Hierarchical RL is an ideal candidate for such endeavors.

The ERD reconstructs the popular notions of an Escape Room as an RL testbed, and
allows RL agents to solve virtual Escape Rooms using the infrastructure we
present in this paper. Each ERD instance consists of a room with a single exit
and a predefined puzzle that the RL agent must solve prior to exiting the room.
The agent's internal state (e.g. joint configurations) and external state (e.g.
world position) are concatenated into a single state vector that can be
manipulated through a set of discrete actions. In the next section we describe
the specific state and action spaces, as well as the transition and reward
functions, which together comprise the MDP of this testbed domain.

\subsecspace
\subsection{The Escape Room MDP}
\subsecspace

Our first step in describing the ERD implementation then is to describe its
theoretical foundation in the language of an MDP. It is important to note that
because the ERD is a ``flexible" domain, it is more accurately described as a
domain \emph{schematic} than a single, static domain in itself. An
\emph{instance} of ERD may be defined based on the specific criteria and
parameters laid out below; any ERD instance is therefore comprised of the
following dimensions:

\begin{enumerate}
  \item The agent's 6-dimensional pose consisting of both 3D position (X, Y,
  and Z) and 3D orientation (Heading, Pitch, and Roll). These state dimensions
  are always continuous. \\
  \item A set of $N_p$ puzzle-specific dimensions that describe the state of
  the puzzle (see Section~\ref{sec:puzzles}) that must be solved in order to unlock the room's exit. These
  state dimensions may be either continuous or discrete. \\
  \item A set of $N_j$ 1-DoF joint positions that describe the joints on the
  agent's virtual robotic arm. The arm may (optionally) be used by the agent to
  interact with the puzzle specific to its ERD instance.
\end{enumerate}

Based on these definitions any ERD instance must have at least 6 state
dimensions, however there are no restrictions on the maximum number of
dimensions. Specific related to the sizes of arm links, orientation of arm
joints, or the transition dynamics of an embedded puzzle are all determined at
the discretion of the researcher who has designed that particular ERD instance.

\begin{figure}[!]
  \centering
  \includegraphics[scale=0.31]{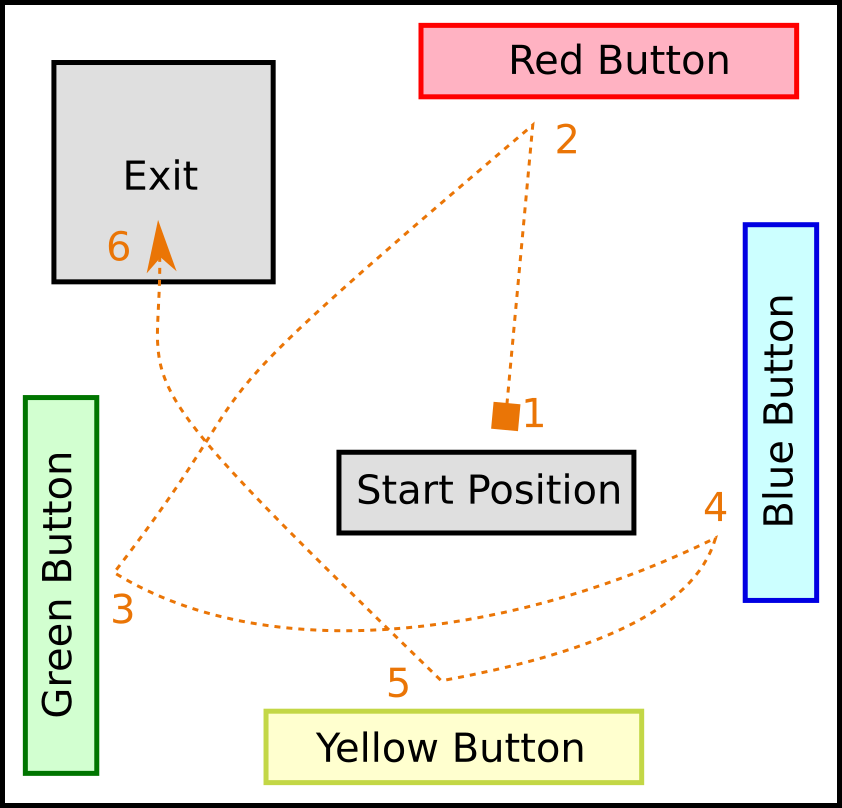}
  \caption{A bird's-eye view of the layout used for the ERD with embedded
    Button Puzzle (see Section~\ref{sec:puzzles}).
  }
  \label{fig:layout}
\end{figure}

The Actions of the ERD are designed to be easily mapped to a real-world robot
agent (e.g. a robot arm) with an auxiliary movement controller, and thus no
explicit actions exist that are tied to a room's embedded puzzle. Instead,
actions only affect the robot's position and its joint positions. The actions
are defined in Table~\ref{tab:actions}.

\begin{table}
  \centering
  \vspace{10pt}
  \def\arraystretch{1.8}
  \begin{tabular}{|c|c|}
    \hline
    {\bf Movement Action} & {\bf Displacement} \\ \hline
    Move Forward/Move Back & 1 Meter \\ \hline
    Strafe Left/Right & 1 Meter \\ \hline
    Turn Left/Right & 10 Degrees \\ \hline
    Increment/Decrement Joint & 10 Degrees \\ \hline
  \end{tabular}
  \vspace{10pt}
  \caption{The ERD action space.}
  \label{tab:actions}
\end{table}

The transition function is governed primarily by the puzzle embedded in each
specific ERD instance, however movement and actuation actions each invoke the
expected state transitions. For example, if the agent executes ``Move Forward",
it will be displaced by approximately 1 meter in whatever direction it is
facing.

The reward function of ERD is intentionally simple. Whenever the agent
takes an action it incurs a reward of -1. When the agent successfully exits the
room, it earns a reward of 100. The purpose of using a sparse reward function
is to encourage general-purpose algorithms that require minimal expert
configuration prior to deployment.

In Section~\ref{sec:experiments} we perform a set of experiments using a small
set of similar ERD instances and provide results that have been averaged over
each. The general layout of these instances is depicted in
Figure~\ref{fig:layout} The path depicted in the figure shows one possible route
that satisfies a hypothetical button dependency configuration. In general, ERD
instances are randomized and the Button Puzzle uses a random seed for generating
its button dependency graph, so the optimal routes change between instances.

\secspace
\section{Software Implementation}\label{sec:puzzles}
\secspace

Our first step in implementing the Escape Room was to pick the software
frameworks that would enable smooth integration with general RL problems, as
well as support the HRL-specific task hierarchies we discussed earlier in this
work. 

The Open AI Gym \cite{gym} is already a popular framework for designing RL test
domains, however there is little support in the way of modeling complex 3D
environments. The Gym's answer to 3D modeling is MuJoCo
\cite{todorov2012mujoco}. While MuJoCo provides much of the infrastructure we
needed for ERD, it is closed-source and relatively expensive. Moreover, MuJoCo's
license restrictions make cluster-based learning prohibitively expensive.
MuJoCo's authors point out that free trials are available, but researchers might
be hesitant to tightly couple their work with services that would eventually be
too expensive to maintain. 

Instead of relying on MuJoCo, we used the Panda3D Game Engine \cite{panda3d}.
Panda3D is a completely free and open source game development framework that
provides all of the basic infrastructure necessary for defining and interacting
with a 3D virtual environment. We know of no other existing integrations between
the Gym and Panda3D, so in creating the ERD we also designed a programmatic
framework for expressing 3D environments as Gym-compatible testing
environments. This framework is included in our ERD GitHub repository.

\subsecspace
\subsection{Puzzles}
\subsecspace

Each ERD instance contains an embedded puzzle that integrates with the physical
makeup of the room. The puzzle can be any physics-based challenge that must be
solved prior to exiting the room. Figure~\ref{fig:layout} shows how we embedded a
Button Puzzle where the buttons must be pressed in a specific order
before the room can be exited. However, this Button Puzzle can be swapped
out for any other puzzle that a researcher (or domain designer) wishes. 
In this way, the puzzle embedded in a given ERD instance dictates the 
overall difficulty of the instance's MDP.

Ideally, each puzzle is further customizable to a minor degree. For
example, in the Button Puzzle the locations and sizes of the buttons can be
adjusted in order to randomize the room and prevent overfitting. One can see
how the puzzle might be extended to add more buttons, or even a variety of
interactive controls to complexify the MDP and provide greater challenge and
randomization to the agent.

\subsecspace
\subsection{Button Puzzle}
\subsecspace

Although the name is new, the Button Puzzle has been used in a variety of HRL
publications under different identities, such as the BitFlip Domain
\cite{diuk2006hierarchical}, the LightBox Domain \cite{Vigorito2010}, the Random
Lights Domain \cite{menashe2015monte}, and Randomly Generated Factored Domains
\cite{hallak2015off}. In each instance, the domain consists of binary variables
that are causally related to one another. Only one variable can be modified at
a time, and so the agent benefits greatly from learning an action hierarchy
that represents the causal structure of the different variables in the domain.
As shown by \cite{menashe2015monte}, complex versions of the Button Puzzle
can be fully modeled and solved by intrinsically-motivated agents in fewer than
20,000 timesteps.

In our implementation there are up to 4 buttons that may be causally related to
one another according to any Directed Acyclic Graph, meaning that a button can
only be switched ``on" after all of its dependent buttons have been switched
``on" as well. Therefore an agent that can deduce the causal structure of the
buttons can quickly determine the correct order in which to toggle them. One
button is designed as the unknown goal, and when this button is toggled the
agent is free to exit the room.

\subsecspace
\subsection{Accessibility}
\subsecspace

The Open AI Gym and the Arcade Learning Environment have shown that reducing the
barrier to entry is a key endeavor to ensure the proliferation of a testing
domain. We have therefore taken every measure to minimize the cost of
integration for ERD. ERD is hosted on our GitHub page\citerepo{}, and can be
integrated with most puzzles so long as they conform to a minimal API. 
We have also ensured that ERD can be run in a number of modes to aid with
the process of debugging integrations or agent performance. The modes are as
follows:

\begin{itemize}
  \item {\bf Manual} - The agent is controlled entirely by the user via
  keyboard presses. \\
  \item {\bf Debug} - The agent is controlled by a provided RL algorithm,
  and a great deal of diagnostic information is reported by the program. \\
  \item {\bf Release} - Time is virtually dilated and visualizations are
  disabled to maximize processing speed. This mode runs approximately 100
  times faster than the other modes, and is optimized for long-term, repeated,
  and unattended experiments.
\end{itemize}

Together, these features make ERD trivial to integrate with existing
Gym-compatible agents for the general RL community. Any agent that is
interoperable with the Open AI Gym can be evaluated against the ERD without
modification.

\secspace
\section{Experiments and Results} \label{sec:experiments}
\secspace

In this section we present the results of a set of experiments that demonstrates
the need for hierarchical action models and algorithms that can integrate such
models into their action selection processes. These experiments show that,
when seemingly trivial problems rely on hierarchical action sequences and sparse
reward functions, these problems can take an extremely long time to solve even
with modern RL algorithms.

The following experiments make use of existing, publicly-available
implementations \cite{plappert2016kerasrl} of the following algorithms:

\vspace{0.35cm}
\begin{enumerate}
  \item Deep Q Learning (DQN) \cite{DBLP:journals/corr/MnihKSGAWR13,mnih2015human} \\
  \item Deep SARSA \cite{zhao2016deep} \\
  \item Deep Deterministic Policy Gradient (DDPG) \cite{DBLP:journals/corr/LillicrapHPHETS15} \\
\end{enumerate}

\begin{figure}[!]
  \centering
  \includegraphics[scale=0.6,trim={14pt 0 0 0},clip]{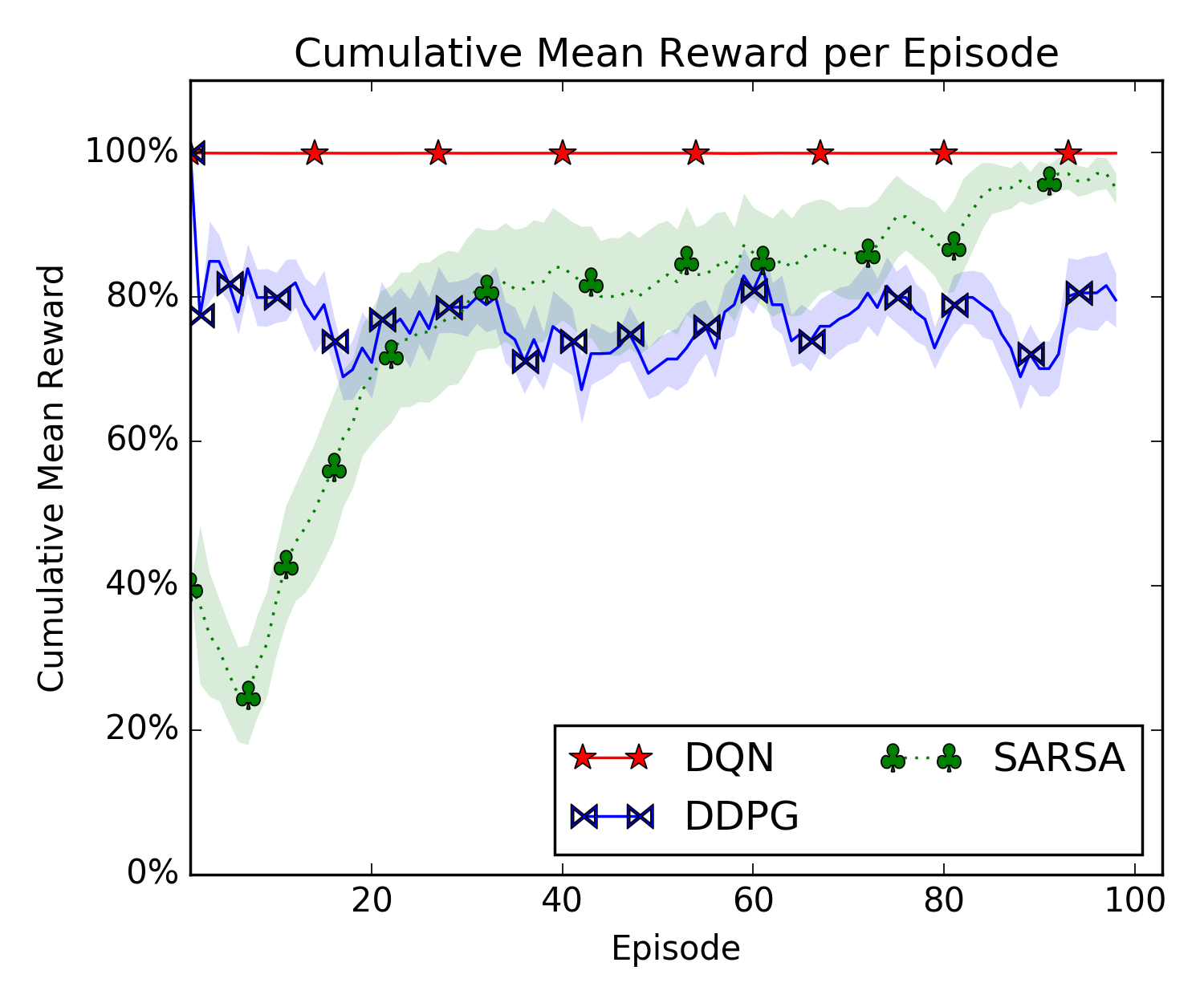}
  \caption{A simple baseline experiment using an ERD instance with one button.
    Each algorithm is able to reach the goal with varying degrees of frequency,
    however DDPG never stabilizes on a single, high-fitness policy.
  }
  \label{fig:results-baseline}
\end{figure}

Figure~\ref{fig:results-baseline} is our baseline experiment which shows that
these algorithms achieve some amount of success on a minimally difficult
instance of ERD, where a single button must be pressed before exiting the
domain. The agent receives a reward of -1 for each timestep, up to a maximum of
1,000 timesteps per episode. If the agent touches all of the buttons and
reaches the goal within 1,000 timesteps, the agent receives an additional
reward of 100. Movement speed is defined such that the goal can be achieved
within approximately 10 timesteps, meaning that the optimal reward is
approximately 90 per episode. The rewards in the figure have been rescaled as
percentages relative to the minimum and maximum cumulative rewards that can be
obtained per episode. Each data series has been averaged over 10 separate trials.

The figure shows that DDPG and DQN reach viable policies almost immediately,
and then repeatedly adjust those policies for the remainder of each trial.
SARSA stabilizes after approximately 100 episodes. Policies that do not earn
maximal reward tend to randomly explore partway through each episode. Small
changes in orientation, for example, can lead to agent trajectories that never
toggle buttons or reach the exit. Each agent consistently exhibited optimal
performance on at least 25\% of episodes by the end of the experiment.

\begin{figure}[!]
  \centering
  \includegraphics[scale=0.6,trim={14pt 0 0 0},clip]{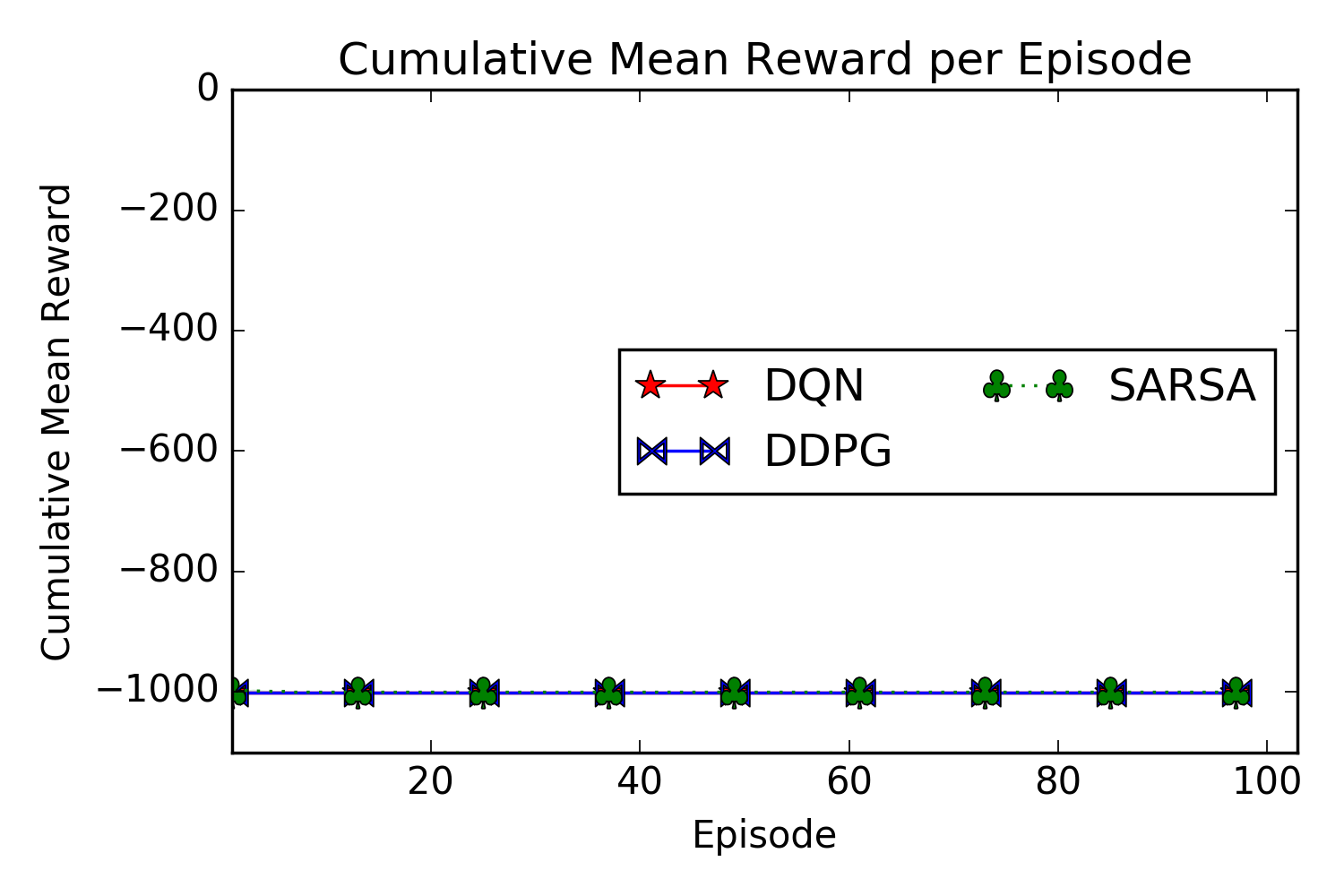}
  \caption{An extension of the baseline experiment using an ERD instance with
    two buttons. None of the tested algorithms were able to reach the goal
    state.
  }
  \label{fig:results-standard}
\end{figure}

In our next experiment we show how the addition of a seemingly trivial
domain element can hamstring these algorithms' ability to quickly learn viable
policies. In Figure~\ref{fig:results-standard}, we have taken the domain from
Figure~\ref{fig:results-baseline} and added a second button, and enforced the
rule that both buttons must be pressed in a particular order before the agent
can reach the domain instance's goal state. Each algorithm was actively trained for
20,000 timesteps. For comparison, other versions of the Button Puzzle found in
earlier work contain 10-20 buttons and can be accurately modeled and solved
within 20,000 timesteps by model-based HRL algorithms
\cite{Vigorito2010,menashe2015monte}.

Figure~\ref{fig:results-standard} demonstrates the drastic reduction in
fitness that is observed for these algorithms when a domain requires
hierarchical action sequences to observe sparse rewards. It is striking that we
see such contrast between deceptively similar domains; however, the fact that
modern Deep RL algorithms fall apart in the face of such challenges reinforces
the claim that a wider variety of testing domains is needed, and in particular,
that domains requiring hierarchical action sequences should be included as
standard benchmarks for forthcoming work.

\vspace{0.5cm}
\subsecspace
\subsection{Meta-Actions}
\subsecspace

In our final experiment we show that the inclusion of a hierarchical
action model is sufficient for solving the proposed multi-button ERD instances
in a timely manner. Direct modification of the algorithms we tested is
outside the scope of this work, so instead we added auto-generated
\emph{meta-actions} that are capable of performing individual button presses
from any state. For example, if an agent is in an ERD instance with 2 buttons, it
will have access to all actions outlined in Table~\ref{tab:actions}, as well as
3 additional actions: \texttt{meta-01}, \texttt{meta-02}, and
\texttt{meta-exit}. These actions construct and execute sequences of primitive
actions that navigate to Button 1, Button 2, and the exit region, respectively.
Each such action appears as a single timestep to the agent, but yields a reward
relative to the number of primitive steps taken (e.g.  -16 for meta-actions that
require 16 primitives).

\begin{figure}[!]
  \centering
  \includegraphics[scale=0.6,trim={14pt 0 0 0},clip]{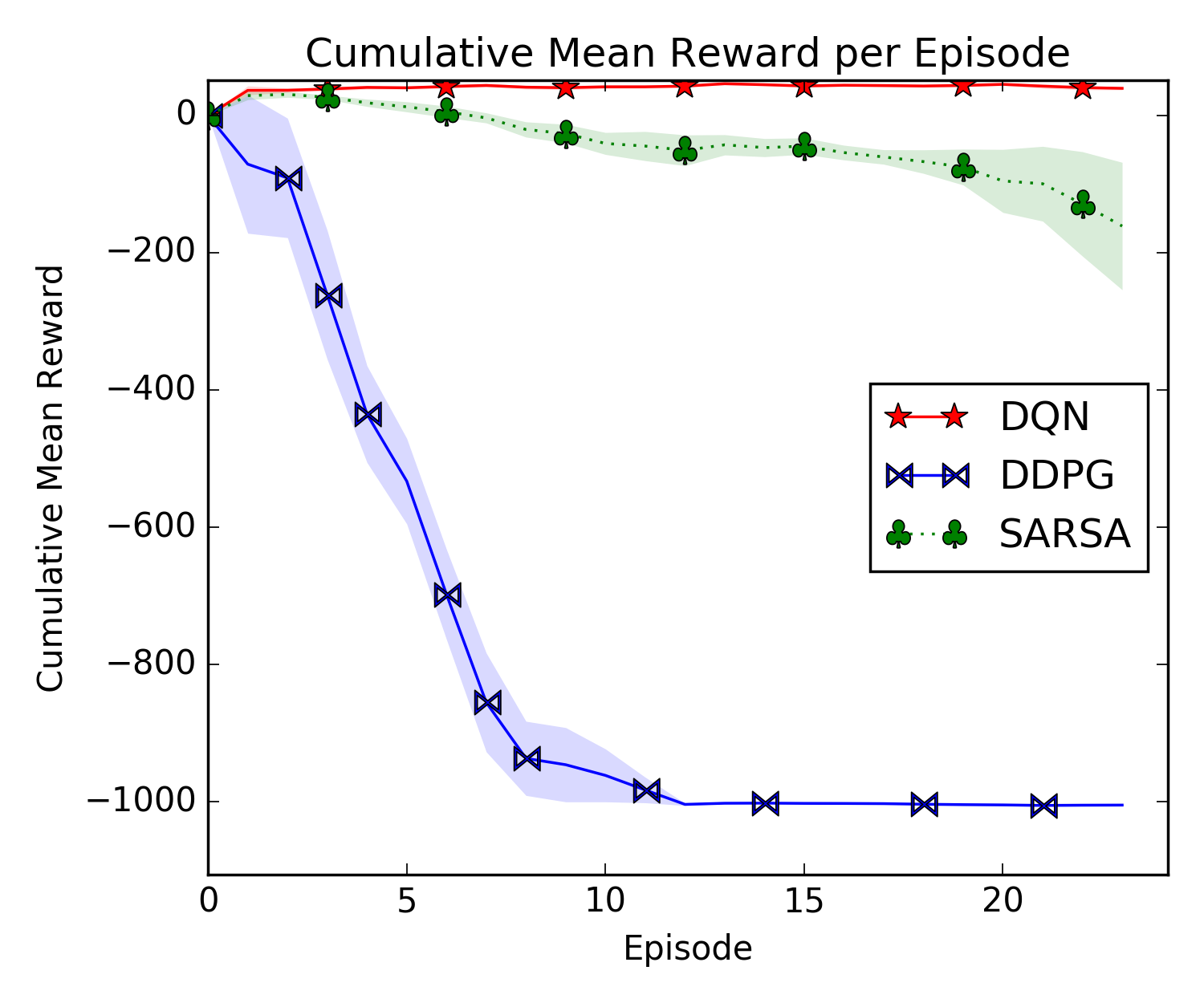}
  \caption{An extension of the baseline experiment using an ERD instance with
    two buttons and additional meta-actions. The meta-actions enabled
    significant performance improvements relative to the primitive-only
    experiment.
  }
  \vspace{0.5cm}
  \label{fig:results-meta}
\end{figure}

The results of the meta-action experiment are shown in
Figure~\ref{fig:results-meta}. These results demonstrate that significant
performance improvements are possible when an accurate action model is
available to an RL agent. However, both SARSA and DDPG show worsening
performance over time. The reason for this phenomenon is that the meta-actions
appear to these agents as exceptionally large state changes that occur over a
single timestep. Because SARSA and DDPG updates consider multiple states and
actions simultaneously, this can negatively impact the accuracy of these
algorithms' deep value networks. DDPG is particularly affected because of its
reliance on computed gradients. While it may be possible to modify these
algorithms to address this issue, these results show that while an action model
can benefit performance, the model must be thoroughly integrated with each
algorithm's action selection and value fitting procedures in order to ensure
increasing fitness over time.

\secspace
\section{Conclusion and Future Work} \label{sec:conclusion}
\secspace
\vspace{0.15cm}

In this paper we have presented the Escape Room Domain, a new testbed domain
designed for advancing HRL research. We have demonstrated that the need for such
testbeds exists, both as a means of bridging the ``difficulty gap" left behind
by existing testbeds, as well as to provide an available, flexible framework for
comparison between a wide array of RL algorithms. The ERD is built on open
source software and is freely available on our public GitHub
repository\citerepo{}.

This domain leaves ample room for future work through collaboration with the RL
community on the GitHub repository. In particular, we plan to use the ERD
as a basis for challenging, evaluating, and comparing new developments in HRL
algorithms. We will be continually updating the repository with references to
new algorithm implementations that make progress toward solving ERD instances
of increasing difficulty.

Alongside our goal of integrating and comparing against new solutions, we will
also be expanding the challenges available via ERD puzzles. The Button Puzzle
we present in Section~\ref{sec:puzzles} is just one example of the possible
subdomains that can be integrated with ERD. Just as the Light Box
Domain \cite{Vigorito2010} has been augmented with a 3D environment to create
the Button Puzzle, we plan to retrofit other classic RL testbeds with the ERD
infrastructure to create new flexible, scalable challenges for RL algorithms.

\clearpage 
\bibliographystyle{plainnat}
\bibliography{paper}

\begin{thebibliography}{55}
\providecommand{\natexlab}[1]{#1}
\providecommand{\url}[1]{\texttt{#1}}
\expandafter\ifx\csname urlstyle\endcsname\relax
  \providecommand{\doi}[1]{doi: #1}\else
  \providecommand{\doi}{doi: \begingroup \urlstyle{rm}\Url}\fi

\bibitem[{Abadi et al.}(2015)]{tensorflow2015-whitepaper}
{Abadi et al.}
\newblock {TensorFlow}: Large-scale machine learning on heterogeneous systems,
  2015.
\newblock URL \url{https://www.tensorflow.org/}.
\newblock Software available from tensorflow.org.

\bibitem[{Association for the Advancement of Artificial
  Intelligence}(2014)]{aaai_website}
{Association for the Advancement of Artificial Intelligence}.
\newblock {Association for the Advancement of Artificial Intelligence}, 2014.
\newblock URL \url{https://www.aaai.org}.

\bibitem[Bacon et~al.(2017)Bacon, Harb, and Precup]{bacon2017option}
Pierre-Luc Bacon, Jean Harb, and Doina Precup.
\newblock The option-critic architecture.
\newblock In \emph{AAAI}, pages 1726--1734, 2017.

\bibitem[{Bellemare} et~al.(2013){Bellemare}, {Naddaf}, {Veness}, and
  {Bowling}]{bellemare13arcade}
M.~G. {Bellemare}, Y.~{Naddaf}, J.~{Veness}, and M.~{Bowling}.
\newblock The arcade learning environment: An evaluation platform for general
  agents.
\newblock \emph{Journal of Artificial Intelligence Research}, 47:\penalty0
  253--279, jun 2013.

\bibitem[Brafman and Tennenholtz(2002)]{brafman2002rmax}
Ronen Brafman and Moshe Tennenholtz.
\newblock R-max - a general polynomial time algorithm for near-optimal
  reinforcement learning.
\newblock \emph{Journal of Machine Learning Research}, 3:\penalty0 213--231,
  2002.

\bibitem[Bratman et~al.(2012)Bratman, Singh, Sorg, and
  Lewis]{bratman2012strong}
Jeshua Bratman, Satinder Singh, Jonathan Sorg, and Richard Lewis.
\newblock Strong mitigation: Nesting search for good policies within search for
  good reward.
\newblock In \emph{Proceedings of the 11th International Conference on
  Autonomous Agents and Multiagent Systems-Volume 1}, pages 407--414.
  International Foundation for Autonomous Agents and Multiagent Systems, 2012.

\bibitem[Brockman et~al.(2016)Brockman, Cheung, Pettersson, Schneider,
  Schulman, Tang, and Zaremba]{gym}
Greg Brockman, Vicki Cheung, Ludwig Pettersson, Jonas Schneider, John Schulman,
  Jie Tang, and Wojciech Zaremba.
\newblock Openai gym, 2016.

\bibitem[{Carnegie Mellon University}(2010)]{panda3d}
{Carnegie Mellon University}.
\newblock Panda3d - free 3d game engine, 2010.
\newblock URL \url{http://panda3d.org}.

\bibitem[Chaganty et~al.(2012)Chaganty, Gaur, and
  Ravindran]{chaganty2012learning}
Arun~Tejasvi Chaganty, Prateek Gaur, and Balaraman Ravindran.
\newblock Learning in a small world.
\newblock In \emph{Proceedings of the 11th International Conference on
  Autonomous Agents and Multiagent Systems-Volume 1}, pages 391--397.
  International Foundation for Autonomous Agents and Multiagent Systems, 2012.

\bibitem[Dayan and Hinton(1993)]{dayan1993feudal}
Peter Dayan and Geoffrey~E Hinton.
\newblock Feudal reinforcement learning.
\newblock In \emph{Advances in neural information processing systems}, pages
  271--278, 1993.

\bibitem[Dean and Kanazawa(1989)]{dean1989model}
Thomas Dean and Keiji Kanazawa.
\newblock A model for reasoning about persistence and causation.
\newblock \emph{Computational intelligence}, 5\penalty0 (2):\penalty0 142--150,
  1989.

\bibitem[Derbinsky et~al.(2012)Derbinsky, Li, and Laird]{derbinsky2012multi}
Nate Derbinsky, Justin Li, and John~E Laird.
\newblock A multi-domain evaluation of scaling in a general episodic memory.
\newblock In \emph{AAAI}, 2012.

\bibitem[Dietterich(2000)]{dietterich2000hierarchical}
Thomas~G Dietterich.
\newblock Hierarchical reinforcement learning with the maxq value function
  decomposition.
\newblock \emph{J. Artif. Intell. Res.(JAIR)}, 13:\penalty0 227--303, 2000.

\bibitem[Diuk et~al.(2006)Diuk, Strehl, and Littman]{diuk2006hierarchical}
Carlos Diuk, Alexander~L Strehl, and Michael~L Littman.
\newblock A hierarchical approach to efficient reinforcement learning in
  deterministic domains.
\newblock In \emph{Proceedings of the fifth international joint conference on
  Autonomous agents and multiagent systems}, pages 313--319. ACM, 2006.

\bibitem[Gundersen and Kjensmo(2017)]{gundersen2017state}
Odd~Erik Gundersen and Sigbj{\o}rn Kjensmo.
\newblock State of the art: Reproducibility in artificial intelligence.
\newblock In \emph{Proceedings of the Thirtieth AAAI Conference on Artificial
  Intelligence and the Twenty-Eighth Innovative Applications of Artificial
  Intelligence Conference}, 2017.

\bibitem[Hallak et~al.(2015)Hallak, Schnitzler, Mann, and
  Mannor]{hallak2015off}
Assaf Hallak, Francois Schnitzler, Timothy Mann, and Shie Mannor.
\newblock Off-policy model-based learning under unknown factored dynamics.
\newblock In \emph{International Conference on Machine Learning}, pages
  711--719, 2015.

\bibitem[Harb et~al.(2017)Harb, Bacon, Klissarov, and Precup]{harb2017waiting}
Jean Harb, Pierre-Luc Bacon, Martin Klissarov, and Doina Precup.
\newblock When waiting is not an option: Learning options with a deliberation
  cost.
\newblock \emph{arXiv preprint arXiv:1709.04571}, 2017.

\bibitem[Hessel et~al.(2017)Hessel, Modayil, Van~Hasselt, Schaul, Ostrovski,
  Dabney, Horgan, Piot, Azar, and Silver]{hessel2017rainbow}
Matteo Hessel, Joseph Modayil, Hado Van~Hasselt, Tom Schaul, Georg Ostrovski,
  Will Dabney, Dan Horgan, Bilal Piot, Mohammad Azar, and David Silver.
\newblock Rainbow: Combining improvements in deep reinforcement learning.
\newblock \emph{arXiv preprint arXiv:1710.02298}, 2017.

\bibitem[Hogg et~al.(2010)Hogg, Kuter, and Munoz-Avila]{hogg2010learning}
Chad Hogg, Ugur Kuter, and Hector Munoz-Avila.
\newblock Learning methods to generate good plans: Integrating htn learning and
  reinforcement learning.
\newblock In \emph{AAAI}, 2010.

\bibitem[{International Foundation for Autonomous Agents and Multiagent
  Systems}(2019)]{aamas_website}
{International Foundation for Autonomous Agents and Multiagent Systems}.
\newblock {International Foundation for Autonomous Agents and Multiagent
  Systems}, 2019.
\newblock URL \url{http://www.ifaamas.org/}.

\bibitem[Jain and Precup(2018)]{jain2018eligibility}
Ayush Jain and Doina Precup.
\newblock Eligibility traces for options.
\newblock In \emph{Proceedings of the 17th International Conference on
  Autonomous Agents and MultiAgent Systems}, pages 1008--1016. International
  Foundation for Autonomous Agents and Multiagent Systems, 2018.

\bibitem[Jong and Stone(2008)]{jong2008hierarchical}
Nicholas~K Jong and Peter Stone.
\newblock Hierarchical model-based reinforcement learning: R-max+ maxq.
\newblock In \emph{Proceedings of the 25th international conference on Machine
  learning}, pages 432--439. ACM, 2008.

\bibitem[Kaelbling(1993)]{kaelbling1993hierarchical}
Leslie~Pack Kaelbling.
\newblock Hierarchical learning in stochastic domains: Preliminary results.
\newblock In \emph{Proceedings of the tenth international conference on machine
  learning}, volume 951, pages 167--173, 1993.

\bibitem[Li et~al.(2016)Li, Narayan, and Leong]{li2016core}
Zhuoru Li, Akshay Narayan, and Tze-Yun Leong.
\newblock A core task abstraction approach to hierarchical reinforcement
  learning.
\newblock In \emph{Proceedings of the 2016 International Conference on
  Autonomous Agents \& Multiagent Systems}, pages 1411--1412. International
  Foundation for Autonomous Agents and Multiagent Systems, 2016.

\bibitem[Li et~al.(2017)Li, Narayan, and Leong]{li2017efficient}
Zhuoru Li, Akshay Narayan, and Tze-Yun Leong.
\newblock An efficient approach to model-based hierarchical reinforcement
  learning.
\newblock In \emph{AAAI}, pages 3583--3589, 2017.

\bibitem[Lillicrap et~al.(2015)Lillicrap, Hunt, Pritzel, Heess, Erez, Tassa,
  Silver, and Wierstra]{DBLP:journals/corr/LillicrapHPHETS15}
Timothy~P. Lillicrap, Jonathan~J. Hunt, Alexander Pritzel, Nicolas Heess, Tom
  Erez, Yuval Tassa, David Silver, and Daan Wierstra.
\newblock Continuous control with deep reinforcement learning.
\newblock \emph{CoRR}, abs/1509.02971, 2015.
\newblock URL \url{http://arxiv.org/abs/1509.02971}.

\bibitem[MacAlpine et~al.(2015)MacAlpine, Depinet, and Stone]{macalpine2015ut}
Patrick MacAlpine, Mike Depinet, and Peter Stone.
\newblock Ut austin villa 2014: Robocup 3d simulation league champion via
  overlapping layered learning.
\newblock In \emph{AAAI}, pages 2842--2848, 2015.

\bibitem[Masson et~al.(2016)Masson, Ranchod, and
  Konidaris]{masson2016reinforcement}
Warwick Masson, Pravesh Ranchod, and George Konidaris.
\newblock Reinforcement learning with parameterized actions.
\newblock In \emph{AAAI}, pages 1934--1940, 2016.

\bibitem[Menashe and Stone(2015)]{menashe2015monte}
Jacob Menashe and Peter Stone.
\newblock {M}onte {C}arlo {H}ierarchical {M}odel {L}earning.
\newblock In \emph{Proceedings of the 2015 International Conference on
  Autonomous Agents and Multiagent Systems}, pages 771--779, 2015.

\bibitem[Mnih et~al.(2013)Mnih, Kavukcuoglu, Silver, Graves, Antonoglou,
  Wierstra, and Riedmiller]{DBLP:journals/corr/MnihKSGAWR13}
Volodymyr Mnih, Koray Kavukcuoglu, David Silver, Alex Graves, Ioannis
  Antonoglou, Daan Wierstra, and Martin~A. Riedmiller.
\newblock Playing atari with deep reinforcement learning.
\newblock \emph{CoRR}, abs/1312.5602, 2013.
\newblock URL \url{http://arxiv.org/abs/1312.5602}.

\bibitem[Mnih et~al.(2015)Mnih, Kavukcuoglu, Silver, Rusu, Veness, Bellemare,
  Graves, Riedmiller, Fidjeland, Ostrovski, et~al.]{mnih2015human}
Volodymyr Mnih, Koray Kavukcuoglu, David Silver, Andrei~A Rusu, Joel Veness,
  Marc~G Bellemare, Alex Graves, Martin Riedmiller, Andreas~K Fidjeland, Georg
  Ostrovski, et~al.
\newblock Human-level control through deep reinforcement learning.
\newblock \emph{Nature}, 518\penalty0 (7540):\penalty0 529, 2015.

\bibitem[Ngo et~al.(2014)Ngo, Ngo, and Wolfgang]{ngo2014monte}
Vien~Anh Ngo, Hung Ngo, and Ertel Wolfgang.
\newblock Monte carlo bayesian hierarchical reinforcement learning.
\newblock In \emph{Proceedings of the 2014 international conference on
  Autonomous agents and multi-agent systems}, pages 1551--1552. International
  Foundation for Autonomous Agents and Multiagent Systems, 2014.

\bibitem[Omidshafiei et~al.(2018)Omidshafiei, Kim, Pazis, and
  How]{omidshafiei2018crossmodal}
Shayegan Omidshafiei, Dong-Ki Kim, Jason Pazis, and Jonathan~P How.
\newblock Crossmodal attentive skill learner.
\newblock In \emph{Proceedings of the 17th International Conference on
  Autonomous Agents and MultiAgent Systems}, pages 139--146. International
  Foundation for Autonomous Agents and Multiagent Systems, 2018.

\bibitem[Osa and Sugiyama(2017)]{osa2017hierarchical}
Takayuki Osa and Masashi Sugiyama.
\newblock Hierarchical policy search via return-weighted density estimation.
\newblock \emph{arXiv preprint arXiv:1711.10173}, 2017.

\bibitem[Osentoski and Mahadevan(2010)]{osentoski2010basis}
Sarah Osentoski and Sridhar Mahadevan.
\newblock Basis function construction for hierarchical reinforcement learning.
\newblock In \emph{Proceedings of the 9th International Conference on
  Autonomous Agents and Multiagent Systems: volume 1-Volume 1}, pages 747--754.
  International Foundation for Autonomous Agents and Multiagent Systems, 2010.

\bibitem[Parr and Russell(1998)]{parr1998reinforcement}
Ronald Parr and Stuart Russell.
\newblock Reinforcement learning with hierarchies of machines.
\newblock \emph{Advances in neural information processing systems}, pages
  1043--1049, 1998.

\bibitem[Plappert(2016)]{plappert2016kerasrl}
Matthias Plappert.
\newblock keras-rl.
\newblock \url{https://github.com/keras-rl/keras-rl}, 2016.

\bibitem[Precup(2000)]{precup2000temporal}
Doina Precup.
\newblock \emph{Temporal abstraction in reinforcement learning}.
\newblock University of Massachusetts Amherst, 2000.

\bibitem[Precup et~al.(1997)Precup, Sutton, and Singh]{precup1997planning}
Doina Precup, Richard~S Sutton, and Satinder~P Singh.
\newblock Planning with closed-loop macro actions.
\newblock In \emph{Working notes of the 1997 AAAI Fall Symposium on
  Model-directed Autonomous Systems}, pages 70--76, 1997.

\bibitem[Roderick et~al.(2018)Roderick, Grimm, and Tellex]{roderick2018deep}
Melrose Roderick, Christopher Grimm, and Stefanie Tellex.
\newblock Deep abstract q-networks.
\newblock In \emph{Proceedings of the 17th International Conference on
  Autonomous Agents and MultiAgent Systems}, pages 131--138. International
  Foundation for Autonomous Agents and Multiagent Systems, 2018.

\bibitem[Ruan et~al.(2015)Ruan, Comanici, Panangaden, and
  Precup]{ruan2015representation}
Sherry~Shanshan Ruan, Gheorghe Comanici, Prakash Panangaden, and Doina Precup.
\newblock Representation discovery for mdps using bisimulation metrics.
\newblock In \emph{AAAI}, pages 3578--3584, 2015.

\bibitem[{Steel~Crate~Games}(2015)]{keeptalking}
{Steel~Crate~Games}.
\newblock Keep talking and nobody explodes.
\newblock Digital Download, 2015.
\newblock URL \url{http://www.keeptalkinggame.com}.

\bibitem[Stodden et~al.(2013)Stodden, Guo, and Ma]{stodden2013toward}
Victoria Stodden, Peixuan Guo, and Zhaokun Ma.
\newblock Toward reproducible computational research: an empirical analysis of
  data and code policy adoption by journals.
\newblock \emph{PloS one}, 8\penalty0 (6):\penalty0 e67111, 2013.

\bibitem[Strehl et~al.(2007)Strehl, Diuk, and Littman]{strehl2007efficient}
Alexander~L Strehl, Carlos Diuk, and Michael~L Littman.
\newblock Efficient structure learning in factored-state {MDP}s.
\newblock In \emph{AAAI}, volume~7, pages 645--650, 2007.

\bibitem[Sullivan and Luke(2012)]{sullivan2012learning}
Keith Sullivan and Sean Luke.
\newblock Learning from demonstration with swarm hierarchies.
\newblock In \emph{Proceedings of the 11th International Conference on
  Autonomous Agents and Multiagent Systems-Volume 1}, pages 197--204.
  International Foundation for Autonomous Agents and Multiagent Systems, 2012.

\bibitem[Taylor(2011)]{taylor2011teaching}
Matthew~E Taylor.
\newblock Teaching reinforcement learning with mario: An argument and case
  study.
\newblock In \emph{Proceedings of the Second Symposium on Educational Advances
  in Artifical Intelligence}, pages 1737--1742, 2011.

\bibitem[Todorov et~al.(2012)Todorov, Erez, and Tassa]{todorov2012mujoco}
Emanuel Todorov, Tom Erez, and Yuval Tassa.
\newblock Mujoco: A physics engine for model-based control.
\newblock In \emph{Intelligent Robots and Systems (IROS), 2012 IEEE/RSJ
  International Conference on}, pages 5026--5033. IEEE, 2012.

\bibitem[{Valve Corporation}(2007)]{portal}
{Valve Corporation}.
\newblock Portal.
\newblock Digital Download, 2007.
\newblock URL \url{https://store.steampowered.com/app/400/Portal/}.

\bibitem[Vezhnevets et~al.(2017)Vezhnevets, Osindero, Schaul, Heess, Jaderberg,
  Silver, and Kavukcuoglu]{vezhnevets2017feudal}
Alexander~Sasha Vezhnevets, Simon Osindero, Tom Schaul, Nicolas Heess, Max
  Jaderberg, David Silver, and Koray Kavukcuoglu.
\newblock Feudal networks for hierarchical reinforcement learning.
\newblock \emph{arXiv preprint arXiv:1703.01161}, 2017.

\bibitem[Vien and Toussaint(2015)]{vien2015hierarchical}
Ngo~Anh Vien and Marc Toussaint.
\newblock Hierarchical monte-carlo planning.
\newblock In \emph{AAAI}, pages 3613--3619, 2015.

\bibitem[Vigorito and Barto(2010)]{Vigorito2010}
Christopher~M. Vigorito and Andrew~G. Barto.
\newblock {Intrinsically motivated hierarchical skill learning in structured
  environments}.
\newblock \emph{IEEE Transactions on Autonomous Mental Development}, 2\penalty0
  (2):\penalty0 132--143, jun 2010.
\newblock ISSN 1943-0604.
\newblock \doi{10.1109/tamd.2010.2050205}.
\newblock URL \url{http://dx.doi.org/10.1109/tamd.2010.2050205}.

\bibitem[Watkins(1989)]{watkins1989learning}
Christopher John Cornish~Hellaby Watkins.
\newblock \emph{Learning from delayed rewards.}
\newblock PhD thesis, University of Cambridge, 1989.

\bibitem[Wiering and Schmidhuber(1997)]{wiering1997hq}
Marco Wiering and J{\"u}rgen Schmidhuber.
\newblock Hq-learning.
\newblock \emph{Adaptive Behavior}, 6\penalty0 (2):\penalty0 219--246, 1997.

\bibitem[Xu and Laird(2010)]{xu2010instance}
Joseph~Z Xu and John~E Laird.
\newblock Instance-based online learning of deterministic relational action
  models.
\newblock In \emph{AAAI}, 2010.

\bibitem[Zhao et~al.(2016)Zhao, Wang, Shao, and Zhu]{zhao2016deep}
Dongbin Zhao, Haitao Wang, Kun Shao, and Yuanheng Zhu.
\newblock Deep reinforcement learning with experience replay based on sarsa.
\newblock In \emph{Computational Intelligence (SSCI), 2016 IEEE Symposium
  Series on}, pages 1--6. IEEE, 2016.

\end{thebibliography}

\end{document}